\documentclass[letterpaper]{article} 
\usepackage{aaai25}  
\usepackage{times}  
\usepackage{helvet}  
\usepackage{courier}  
\usepackage[hyphens]{url}  
\usepackage{graphicx} 
\usepackage{natbib}  
\usepackage{caption} 
\usepackage{algorithm}
\usepackage{algorithmic}
\usepackage[percent]{overpic}
\usepackage{lipsum} 
\usepackage{booktabs} 
\usepackage{amsthm,amsmath,amssymb}
\usepackage{multirow}
\usepackage{bbding}
\usepackage{xcolor}

\newcommand{\myparagraph}[1]{\vspace{0.0em}\noindent\vspace{0.0em}\textbf{#1}}
\usepackage{newfloat}
\usepackage{listings}
\DeclareCaptionStyle{ruled}{labelfont=normalfont,labelsep=colon,strut=off} 
\floatstyle{ruled}
\newfloat{listing}{tb}{lst}{}
\floatname{listing}{Listing}
\title{3CAD: A Large-Scale Real-World 3C Product Dataset for Unsupervised Anomaly Detection}

\author{
    Enquan Yang\textsuperscript{\rm 1}\thanks{These authors contributed equally.},
    Peng Xing\textsuperscript{\rm 2}\footnotemark[1],
    Hanyang Sun\textsuperscript{\rm 1},
    Wenbo Guo\textsuperscript{\rm 1},
    Yuanwei Ma\textsuperscript{\rm 3},
    Zechao Li\textsuperscript{\rm 2},
    Dan Zeng\textsuperscript{\rm 1}\thanks{Corresponding author}
}
\affiliations{
    \textsuperscript{\rm 1}School of Communication and Information Engineering, Shanghai University\\
    \textsuperscript{\rm 2}Nanjing University of Science and Technology\\
    \textsuperscript{\rm 3}Changzhou Microintelligence Corporation\\

    \{enquanyang,sunhanyang,guowenbo,dzeng\}@shu.edu.cn, \{xingp\_ng,zechao.li\}@njust.edu.cn, yuanweima@gmail.com
}

\usepackage{bibentry}

\begin{document}

\maketitle

\begin{abstract}
Industrial anomaly detection achieves progress thanks to datasets such as MVTec-AD and VisA. However, they suffer from limitations in terms of the number of defect samples, types of defects, and availability of real-world scenes. These constraints inhibit researchers from further exploring the performance of industrial detection with higher accuracy.
To this end, we propose a new large-scale anomaly detection dataset called 3CAD, which is derived from real 3C production lines.
Specifically, the proposed 3CAD includes eight different types of manufactured parts, totaling 27,039 high-resolution images labeled with pixel-level anomalies. 
The key features of 3CAD are that it covers anomalous regions of different sizes, multiple anomaly types, and the possibility of multiple anomalous regions and multiple anomaly types per anomaly image.
This is the largest and first anomaly detection dataset dedicated to 3C product quality control for community exploration and development.
Meanwhile, we introduce a simple yet effective framework for unsupervised anomaly detection: a \textbf{C}oarse-to-\textbf{F}ine detection paradigm with \textbf{R}ecovery \textbf{G}uidance (CFRG). 
To detect small defect anomalies, the proposed CFRG utilizes a coarse-to-fine detection paradigm. Specifically, we utilize a heterogeneous distillation model for coarse localization and then fine localization through a segmentation model. In addition, to better capture normal patterns, we introduce recovery features as guidance. 
Finally,  we report the results of our CFRG framework and popular anomaly detection methods on the
3CAD dataset, demonstrating strong competitiveness and providing a highly challenging benchmark to promote the development of the anomaly detection field.
Data and code are available: \url{https://github.com/EnquanYang2022/3CAD.}
\end{abstract}

\begin{figure}
    \centering
    \includegraphics[width=.99\linewidth]{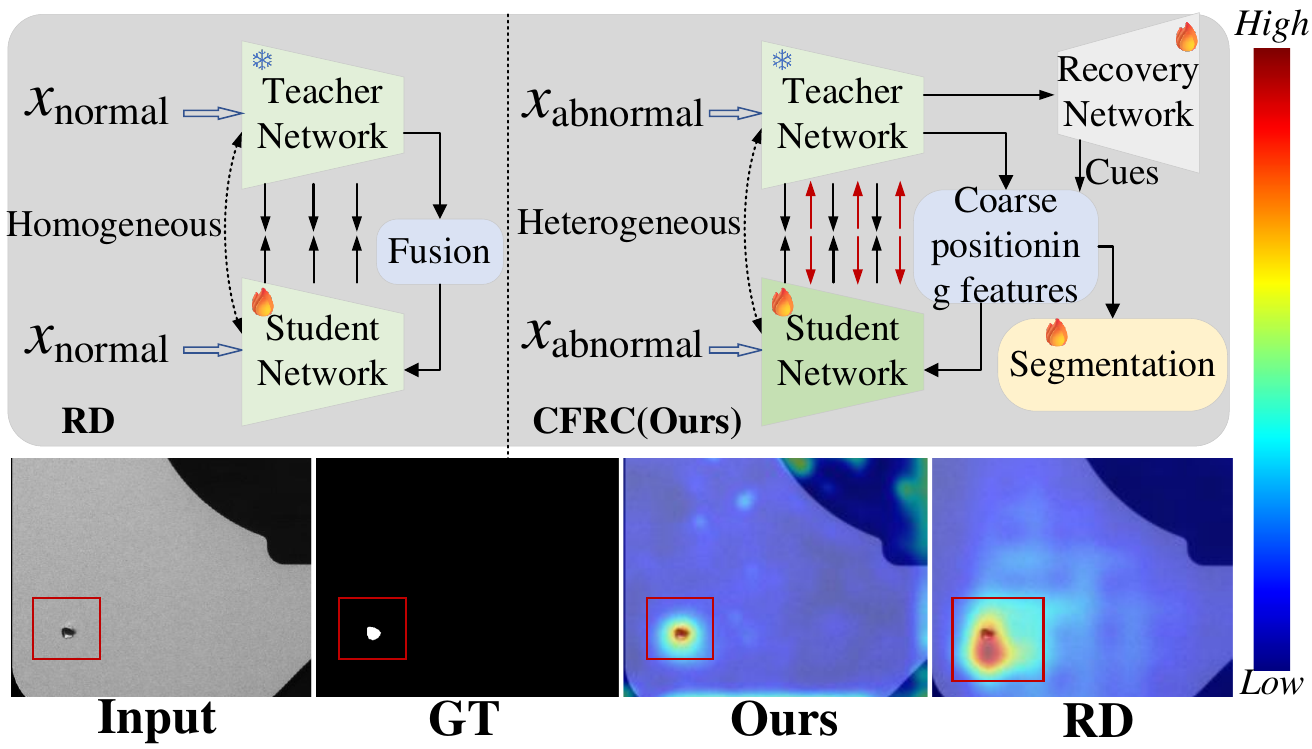}
    \caption{Comparison of previous anomaly detection distillation paradigm with our paradigm. First row: Left: reverse distillation; Right: our proposed paradigm.}
\vspace{-0.3cm}
\label{fig1}
\end{figure}

\section{Introduction}
The rapid expansion of 3C product manufacturing has surpassed the capabilities of traditional manual quality inspection, underscoring the need for advanced algorithms like Image Anomaly Detection (IAD) \cite{(1)liu2024deep}.
Deep learning approaches have demonstrated significant effectiveness in identifying complex and subtle defects in industrial images, enhancing both detection accuracy and robustness \cite{(2)cao2024survey, (3)zhang2023prototypical, (4)gu2024rethinking}.

Typically, supervised methods are employed when the types of defects are known and a sufficient amount of labeled data is available.
These methods are based on classification \cite{(5)nagata2019fusion}, detection \cite{(6)zhang2020concrete}, and segmentation models \cite{(7)zou2018deepcrack}. 
However, obtaining defective samples and enough labeled data can be challenging in practice. 
As a result, unsupervised deep learning methods, which only require normal samples for training, have gained increasing attention \cite{(8)roth2022towards,(9)liu2023simplenet,(39)cao2023collaborative,(10)liu2024unsupervised}.

The introduction of widely-used industrial datasets such as MVTec-AD \cite{(11)bergmann2019mvtec}, VisA \cite{(12)zou2022spot}, and MPDD \cite{(13)jezek2021deep}, has spurred significant academic interest in anomaly detection and led to numerous innovative approaches \cite{(40)xing2023normal,(41)xing2024adps}. However, as research advances, several limitations of these datasets have become apparent.
Firstly, the number of anomaly data poses a challenge. For instance, some categories of anomaly images in MVTec-AD consist of fewer than 60 samples, which may not adequately represent the performance of algorithms in practical applications.
Secondly, there is a concern regarding the authenticity of anomalies. The defects in MVTec-AD, VisA, and Real-IAD \cite{(14)wang2024real} are primarily artificially created, such as scratches or deletions, despite being derived from real products. This creates a gap between the types of anomalies present in these datasets and those encountered in real-world production lines.
Lastly, there is a tendency toward saturation in performance metrics. Recent methods achieve over 99\% in both I-AUROC (image-level) and P-AUROC (pixel-level) metrics, making it challenging to discern the relative merits of new approaches.
Additionally, there is currently no comprehensive anomaly detection dataset tailored specifically for 3C products, further highlighting a gap in the available resources.

To address the limitations of existing datasets and provide data closer to real scenes, we propose a new industrial anomaly detection dataset, named 3CAD, which is derived from real production lines. This dataset focuses on 3C product parts and offers the following advantages:
\begin{itemize}
    \item Real-world Relevance. The dataset includes defects generated during the manufacturing process, reflecting real production scenarios with three common materials and eight different defect types.
    \item Large Scale. With 27,039 high-resolution images, it surpasses most existing anomaly detection datasets.
    \item Varied Defect Distribution.  A single image may contain one or multiple defects of the same or different types, depending on the manufacturing process. The defect location is random.
    \item Complex Defect Morphology. Defects vary significantly in shape, size, and appearance, with many are similar to normal product features.
    \item Challenging Detection. Tiny and hidden defects make accurate detection difficult.
\end{itemize}
We assess several unsupervised anomaly detection algorithms using the 3CAD dataset. The results indicate that, although existing methods perform effectively on popular datasets, they encounter difficulties with precise defect localization in our dataset. 
This suggests there is significant potential for further improvement, as illustrated in Fig.~\ref{fig1}.

To address the challenges of the 3CAD dataset, we propose a \textbf{C}oarse-to-\textbf{F}ine localization paradigm with \textbf{R}ecovery \textbf{G}uidance (CFRG), which enhances distillation and segmentation techniques by integrating a recovery task. CFRG consists of:  
Heterogeneous Feature Extraction: Teacher-student networks with different architectures extract diverse features from the same data, effectively pinpointing abnormal areas and addressing feature redundancy, especially for small and subtle defects.  
Recovery Network: Restores normal features from abnormal ones, capturing the underlying patterns of normal images.  
Segmentation Fine Localization: Leverages the recovery network’s weights to guide the weighting of abnormal features extracted through distillation, which are then input into the segmentation network to improve localization accuracy.  
We hope this work advances anomaly detection in 3C products and inspires further research.

\section{Related Work}

\myparagraph{Anomaly Detection Datasets}: Datasets are crucial for defect detection research. Traditionally, algorithms are developed using specialized datasets for specific objects, such as pcbs \cite{(15)tang2019online}, tiles \cite{(16)huang2020surface}, and steel \cite{(17)he2019end}. These training datasets often required manual labeling, limiting their impact on advancing industrial anomaly detection (IAD). The release of MVTec-AD in 2019 is a significant milestone, as it supported the development of unsupervised IAD algorithms by providing a diverse dataset. Subsequently, datasets like BTAD \cite{(18)mishra2021vt}, MPDD \cite{(19)jezek2021deep}, and VisA \cite{(12)zou2022spot} have further propelled IAD research. Recently, the Real-IAD \cite{(14)wang2024real} dataset introduced a larger, multi-view dataset, presenting new challenges for IAD. 

Anomaly synthetic: Artificially synthesizing anomalies allows researchers to augment datasets and improve model performance, even with limited original data. Recent unsupervised anomaly detection methods have increasingly utilized synthetic anomaly images. For example, CutPaste \cite{(21)li2021cutpaste} generates negative samples by cutting and pasting image segments, while DRAEM \cite{(22)zavrtanik2021draem} and DeSTSeg \cite{(23)zhang2023destseg} use Perlin noise and random uniform samples to create anomaly masks. Additionally, DTD \cite{(24)cimpoi2014describing} provides a source for blending anomalies into original images. The capabilities of diffusion models have further expanded synthetic data generation \cite{(25)hu2024anomalydiffusion, (26)zhang2024realnet}. However, synthetic approaches may yield unrealistic anomalies, and their diversity is limited by the inherent cognitive scope of the model.

\begin{figure*}[t]
    \centering
    \includegraphics[width=.99\linewidth]{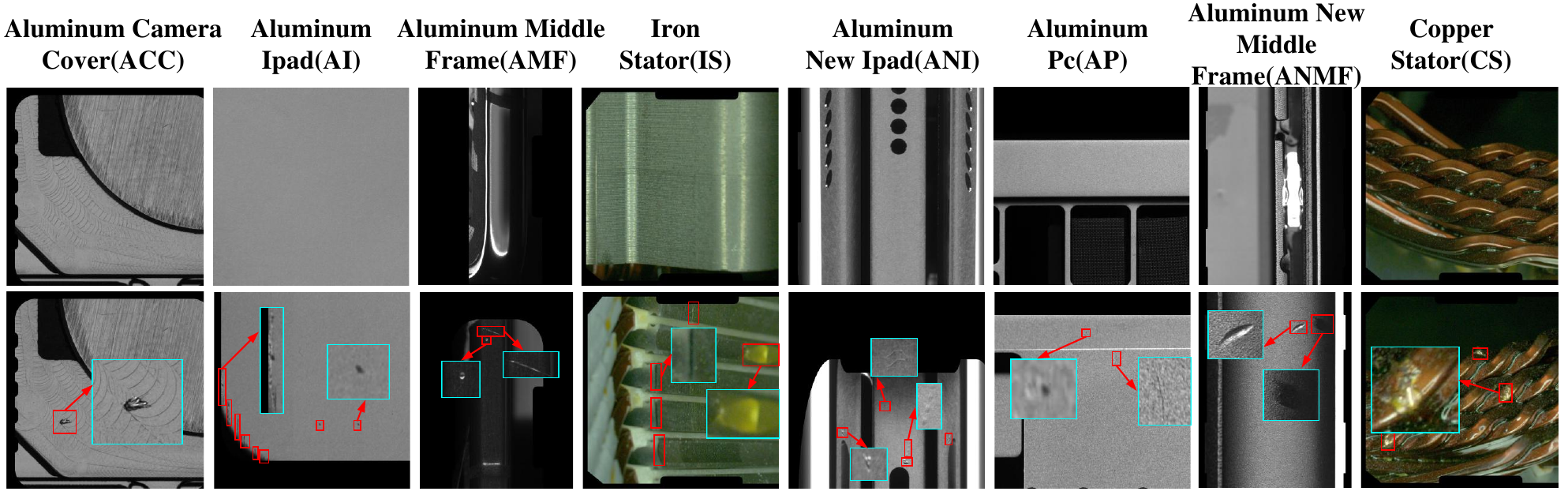}
    \caption{3CAD dataset samples. The first row shows normal images, while the second row displays defective images.}
\label{fig2}
\end{figure*}

\myparagraph{Anomaly Detection}
Recovery-based methods train networks to restore defects in images to their normal state \cite{(27)zavrtanik2021reconstruction, (28)xing2023visual,(44)xing2024recover}. For example, RealNet \cite{(26)zhang2024realnet} employs a feature reconstruction network with pre-trained multi-scale features, adaptively selecting and reconstructing residuals. By avoiding equal inputs and outputs, these methods mitigate the identity mapping issue in traditional reconstruction approaches. Moreover, the adaptability of diffusion models to various downstream tasks has spurred advancements in anomaly detection \cite{(42)shen2023advancing,(43)he2024diffusion}.

Knowledge distillation (KD) methods align teacher-student outputs for normal regions while differentiating defective ones for precise localization \cite{(29)salehi2021multiresolution}. 
Identical network structures, however, may reduce feature diversity. Techniques like RD \cite{(30)deng2022anomaly} and AST \cite{(31)rudolph2023asymmetric} address this by adopting serial or asymmetric architectures, improving differentiation between normal and abnormal features. Our method enhances this further by using heterogeneous teacher-student networks to better separate normal and abnormal features.

\section{The 3CAD Dataset}

\begin{table*}[h]
\centering
\small\renewcommand{\arraystretch}{1.0}
\begin{tabular*}{\linewidth}{@{\extracolsep{\fill}}l|ccccccccc}
\hline
Category &
  \begin{tabular}[c]{@{}c@{}}Training\\ Images\end{tabular} &
  \begin{tabular}[c]{@{}c@{}}Test\\ Images(all)\end{tabular} &
  \begin{tabular}[c]{@{}c@{}}Test\\ Images(good)\end{tabular} &
  \begin{tabular}[c]{@{}c@{}}Test\\ Images(defect)\end{tabular} &
  \begin{tabular}[c]{@{}c@{}}Defect\\ types\end{tabular} &
  \begin{tabular}[c]{@{}c@{}}Image\\ Height\end{tabular} &
  \begin{tabular}[c]{@{}c@{}}Image\\ Width\end{tabular} &
  \begin{tabular}[c]{@{}c@{}}NE\\ / TE\end{tabular} \\ \hline
  ACC      & 784   & 1446  & 369  & 1077  & 10 & 288$\sim$1024  & 288$\sim$1024  & 1$\sim$6/1$\sim$1  \\
AI               & 2096  & 2047  & 913  & 1134  & 3  & 760$\sim$1024  & 600$\sim$1024  & 1$\sim$10/1$\sim$2 \\
AMF       & 1548  & 1479  & 731  & 748   & 5  & 540$\sim$1024  & 800$\sim$950   & 1$\sim$9/1$\sim$4  \\
ANMF & 1072  & 1406  & 670  & 736   & 6  & 400$\sim$1024  & 430$\sim$1024  & 1$\sim$6/1$\sim$2  \\
ANI          & 2233  & 4936  & 999  & 3937  & 4  & 420$\sim$1024  & 580$\sim$1024  & 1$\sim$23/1$\sim$2 \\
AP                 & 1698  & 3161  & 911  & 2250  & 14 & 430$\sim$1024  & 409$\sim$1024  & 1$\sim$12/1$\sim$3 \\
CS               & 409   & 959   & 196  & 763   & 1  & 1024$\sim$1024 & 1024$\sim$1024 & 1$\sim$9/1$\sim$1  \\
IS                 & 653   & 1112  & 295  & 817   & 4  & 1024$\sim$1024 & 1024$\sim$1024 & 1$\sim$12/1$\sim$2 \\
All                          & 10493 & 16546 & 5084 & 11462 & 47 &  -             & -              & -                   \\  \hline
\end{tabular*}
\caption{ 
Statistical overview of the 3CAD dataset. The NE and TE in the last column indicate the number of anomalous regions and the number of anomalous types present in each defective image, respectively.
}
\vspace{-0.2cm}
\label{T1}
\end{table*}

Developing new datasets is crucial for maintaining high quality in 3C products and enhancing the effectiveness of anomaly detection for complex anomalies. 
Given the growing importance of unsupervised anomaly detection in industrial inspections, we introduce the 3CAD dataset, tailored for real-world 3C manufacturing scenarios.

\subsection{Data Construction}

\subsubsection{Data Source.}
The data originates from high-quality segmentation datasets accumulated by the company over several years from various production line projects.
It is specifically tailored for defect detection in 3C products within industrial production processes.

\subsubsection{Data Collection, Annotation.}

The data acquisition process consists of two stages. In the first stage, a series of dedicated mechanisms are employed, including loading, inspection, model analysis, and unloading. 
The workpiece is introduced into this automated system, where images of each part are captured. These images are then analyzed by a model to detect defects.
During the commissioning stage, specialized quality control staff collect both defective and non-defective materials from the production line and photograph them using the designated equipment.
Subsequently, labeling staff annotate the images with pixel-level precision based on the assessments of Quality Engineers (QEs), using a self-developed labeling software similar to LabelMe.

\begin{figure}[t]
    \centering
    \includegraphics[width=\linewidth]{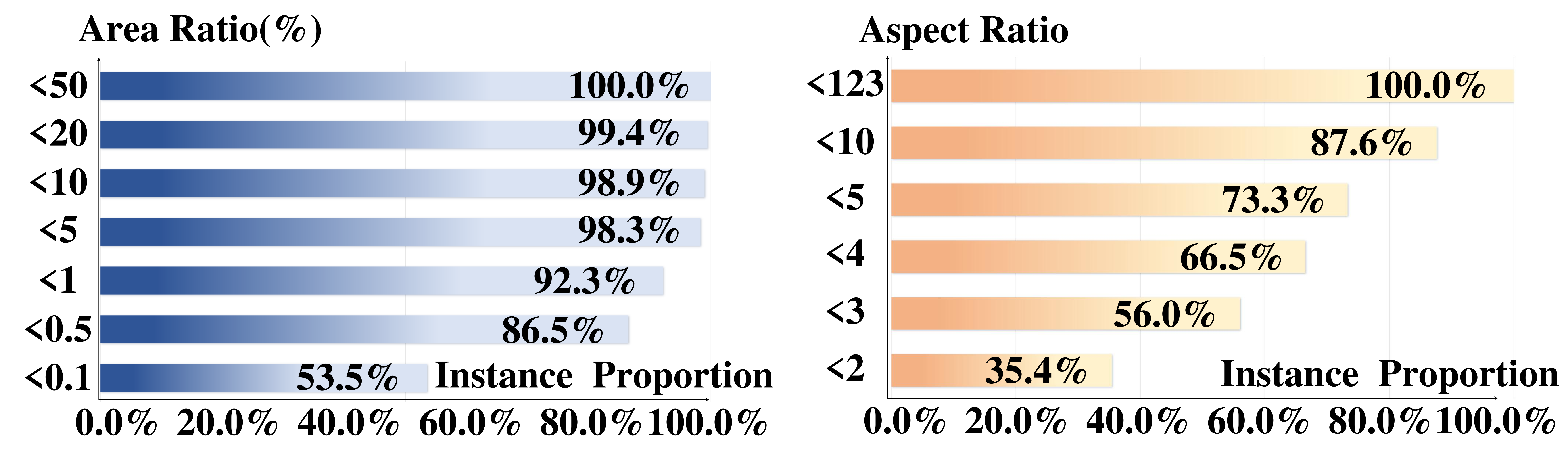}
    \makebox[0.208\textwidth]{\footnotesize a}
    \makebox[0.208\textwidth]{\footnotesize b}
    \caption{Statistics of the proposed 3CAD dataset: a) Defect area ratio. b) Aspect ratio of the minimum bounding rectangle for the defect area.}
    \vspace{-0.3cm}
    \label{fig3}
\end{figure}

\subsubsection{Data Cleaning.}

Data cleaning and labeling are seamlessly integrated to ensure high-quality results. The algorithm autonomously handles certain cleaning tasks, such as consolidating data types based on its schema. To maintain high standards in labeling, annotators undergo daily evaluations to ensure their understanding of defect morphology is accurate and consistent.

Additionally, a review system is in place where the labeling team leader assesses the work of annotators. If any inconsistencies or quality issues are detected, the data is sent back for re-labeling. Typically, under standard project conditions and without interruptions, each annotator processes an average of 10 to 50 samples per day.
For details on dataset construction, see the supplementary material.

\subsubsection{Dataset Description.}

The details of the proposed 3CAD dataset are shown in Tab.~\ref{T1}.
(1) It comprises 10,493 training images and 16,546 testing images that are carefully selected to represent the best acquisition for each product type.
Each defect image is labeled with high-quality pixel-level defect labels.
(2) 3CAD covers a wide range of 3C products from real-world product lines, including complex items like the Aluminum Camera and simple items like Ipad product. 
(3) Rich anomaly categories. Defect types in the dataset range from prominent ones like bumps and dents to more subtle ones such as scratches and pinholes. Meanwhile, a single defect image may simultaneously contain multiple types or multiple abnormal regions (as shown in Fig.~\ref{fig2}).
(4) Diverse anomalous regions. As shown in Fig.~\ref{fig3}.a, 3CAD covers both large-scale anomalous regions and small-scale anomalous areas, with a greater proportion of the currently challenging small-scale anomalies.  Fig~\ref{fig3}.b depicts the distribution of the number of minimum outer rectangular aspect ratios of the anomalous regions, showing the diversity of the morphological distribution of the anomalous regions in 3CAD.

\begin{table}[t]
\centering
\small\renewcommand{\arraystretch}{1.0}

\resizebox{1\linewidth}{!}{
\begin{tabular}{@{}lccccc@{}}
\hline
Dataset  & Normal & Abnormal & \begin{tabular}[c]{@{}l@{}}Defect Source\end{tabular} & \begin{tabular}[c]{@{}l@{}}MDI\end{tabular}  & \begin{tabular}[c]{@{}l@{}}MDT\end{tabular} \\ \hline
BTAD     & 2250   & 580                       & Real                         & \XSolidBrush                               & \XSolidBrush                           \\
MPDD     & 1064   & 282                       & Real                          & \Checkmark                               & \XSolidBrush                           \\
MVTec-AD & 4096   & 1258                    & Forged                           & \XSolidBrush                               & \XSolidBrush                           \\
VisA     & 10621  & 1200                   & Forged                           & \Checkmark                               & \XSolidBrush                           \\
Real-IAD & 99721  & 51329                  & Forged                          & \XSolidBrush                               & \XSolidBrush                           \\
Ours     & 15577  & 11462                    &Real                          & \Checkmark                                & \Checkmark                                            \\ \hline
\end{tabular}%
}
\caption{ 
Comparison with the current popular IAD datasets. \Checkmark: Satisfied. \XSolidBrush: Not satisfied. MDI: Multiple Defect Instances in One Image; MDT: Multiple Defect Types in One Image. 
}
\label{T2}
\end{table}

\subsubsection{Comparison with Popular Datasets.}
Tab.~\ref{T2} shows the results of comparing the proposed 3CAD with other popular datasets. Firstly, 3CAD is derived directly from real manufacturing environments, capturing diverse scenarios where the same defects can exhibit significant variation and different defects can appear quite similar. 
In contrast, datasets such as MVTec-AD, VisA, and Real-IAD tend to rely on artificially generated defect types and distributions that are not conducive to real-world applications. 
The authenticity of 3CAD ensures that algorithms trained on it are better suited for real-world deployment. Moreover, 3CAD includes 15,577 normal samples and 11,462 abnormal samples, surpassing BTAD and MPDD in both scale and variety of defect types.
Last but not least, 3CAD is unique in its may inclusion of multiple defect instances and types in single image. While other datasets like MPDD and VisA allow for multiple instances, they contain a limited number of defect instances and they fall short in presenting multiple defect types within a single image. This feature of 3CAD is particularly important as it mirrors real-world scenarios where products may have more defect instances and more than one type of defect occurring simultaneously, challenging the detection algorithms to identify each one accurately. 

\section{Method}
\begin{figure*}
    \centering
    \centering
    \includegraphics[width=.99\linewidth]{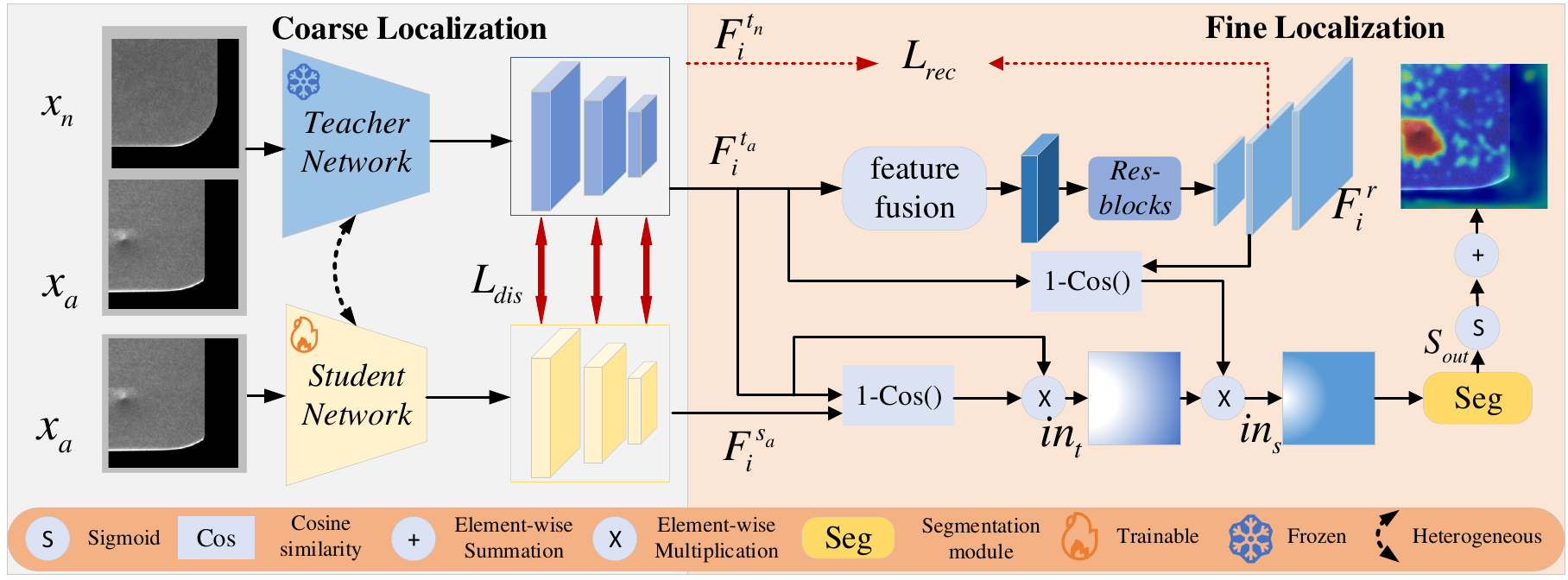}
   
    \caption{The proposed CFRG framework comprises two components: 1) a distilled localization network and 2) a refined segmentation network with restored hints. During training, in the first stage, \(x_a\) and \(x_n\) are input into the teacher network, while \(x_a\) is input into the student network, and the distillation loss between the teacher and the student is calculated. In the second stage, the teacher's features are weighted using the first-stage localization weights and the recovery branch's hint weights, then input into the segmentation network. During testing, the recovery branch generates the localization result from the input and \(\{F_{i}^{r}\}_{i=1}^{K}\), which is then added to the output \(S_{out}\) of the segmentation network to obtain the final anomaly map.}
\label{fig4}
\end{figure*}

To tackle the challenges posed by 3CAD, we propose a simple but effective framework, \textbf{C}oarse-to-\textbf{F}ine detection paradigm with \textbf{R}ecovery \textbf{G}uidance, called CFRG, whose framework is shown in Fig.~\ref{fig4}.
To mitigate the challenges posed by small defects in 3CAD and the discrepancy between real and synthetic anomalies, we propose to use distillation for initial coarse localization, followed by guidance through an anomaly recovery task, and finally fine localization of the anomalies through a segmentation network.

\subsection{Knowledge Distillation for Coarse Localization}
To localize anomalous regions, we use knowledge distillation with a pre-trained teacher network (WideResNet50 \cite{(32)zagoruyko2016wide}) and a learnable student network  (EfficientNet-b0 \cite{(33)tan2019efficientnet}). The student mimics the behavior of the teacher on normal samples and learns to distinguish anomalies, reducing mislocalization, especially for subtle defects. The heterogeneous design, where the teacher and student focus on different aspects of feature extraction, enhances the detection of small, weak, and background-like anomalies. 

Given an input image \(x_n\in\mathbb{R}^{C\times H \times W} \), anomalies are synthesized using DTD and 2D Perlin noise to produce \(x_a\). Multi-level features \(\{F_{i}^{t_a}\}_{i=1}^{K}\in\mathbb{R}^{C_{i}\times H_{i}\times W_{i}} \) and \(\{F_{i}^{s_a}\}_{i=1}^{K}\in\mathbb{R}^{C_{i}\times H_{i}\times W_{i}} \) are extracted from \( x_a \) by teacher and student networks. We start by calculating the cosine similarity at each stage:

\begin{equation}
\mathcal{L}_{i}^{\mathrm{cos}}\left(x,y\right)=1-\frac{F_{i}^{t_a}\left(x,y\right)}{\|F_{i}^{t_a}\left(x,y\right)\|_2}\cdot\frac{F_{i}^{S_a}\left(x,y\right)}{\|F_{i}^{S_a}\left(x,y\right)\|_2},
\label{EN3}
\end{equation}
 where indices \(i\) and \(j\) denote the spatial coordinates on the feature map. Next, using the ground-truth mask to differentiate between normal and abnormal regions, we construct a loss function that maximizes the cosine similarity within normal regions while minimizing it in abnormal regions:
 \begin{equation}
L_{dis}=\sum_{i=1}^3\left(\left(1-G\right)\mathcal{L}_{i}^{\mathrm{cos}}\left(x,y\right)+G\left(1-\mathcal{L}_{i}^{\mathrm{cos}}\left(x,y\right)\right)\right),
\label{EN4}
\end{equation}
where \( G \in \{0, 1\} \), indicating the Ground Truth mask.

\subsection{Recovery Feature as Guidance}
Recovery-based methods train networks to treat anomalies as noise and focus on reconstructing images to their normal state. This approach helps the network learn and represent the intrinsic patterns of normal images, reducing sensitivity to real-world anomalies and mitigating distillation localization bias, even when simulated and real anomalies differ. In our framework, a pre-trained teacher network extracts multi-scale features \(\{F_{i}^{t_n}\}_{i=1}^{K}\) and \(\{F_{i}^{t_a}\}_{i=1}^{K}\) from normal and abnormal images, respectively. These features capture both global and local details, enhancing model robustness to noise. Following RD \cite{(30)deng2022anomaly}, we fuse \(\{F_{i}^{t_a}\}_{i=1}^{K}\) into the dimension of the final layer and input it into a recovery network. This network, using a ResNet block with transposed convolution, generates multi-layer features \(\{F_{i}^{r}\}_{i=1}^{K}\) that match the dimensions of \(\{F_{i}^{t_a}\}_{i=1}^{K}\).

 \begin{equation}
\{F_i^r\}_{i=1}^K=Resblock\left(Bn\left(F_{i}^{t_{a}}\right)\right),
\label{EN5}
\end{equation}
where \(Bn\) denotes the feature fusion operation, while \(Resblock\) refers to the stacked ResNet blocks. Our goal is to align the feature spaces of\(\{F_{i}^{t_n}\}_{i=1}^{K} \) and\(\{F_{i}^{r}\}_{i=1}^{K} \) to optimize the recovery network. We still use cosine similarity loss to achieve this to obtain \(L_{rec}\).

\subsection{Segmentation for Fine-Grained Localization}

Pixel-level localization by distillation and recovery network works well for clear anomalies but is insufficient for more complex cases in our data. To improve precision, we integrate segmentation techniques.

First, we compute the cosine similarity \(w_d\) between \(\{F_{i}^{t_a}\}_{i=1}^{K}\) and \(\{F_{i}^{s_a}\}_{i=1}^{K}\), where lower values indicate abnormal regions. The anomaly localization is represented by \(1-w_d\), applied to the features of teacher network to obtain \(in_t\). A similar process is applied to the input and output of recovery network, yielding \(1-w_r\) to refine \(in_t\) and produce \(in_s\). Finally, \(in_s\) is input into a segmentation module, which fuses multi-layer information through a skip connection and outputs the segmentation mask \(S_{out}\) in the original image size. The segmentation module is optimized using binary cross-entropy loss.

\begin{equation}
\begin{aligned}L_{bce}&=\frac{1}{H\times W}\sum_{x=1}\sum_{y=1}[-G(x,y)\log S_{out}(x,y)\\&- (1-G(x,y)\log(1-S_{out}(x,y))],\end{aligned}
\label{EN8}
\end{equation}
Total loss of CFRG is expressed as:

 \begin{equation}
L_{all}=L_{dis}+L_{rec}+L_{bce}
\label{EN9}
\end{equation}

\subsection{Model Inference}
During the inference phase, we measure the cosine similarity between the output of teacher and\(\{F_{i}^{r}\}_{i=1}^{K} \). This result is then combined with the segmentation mask \( S_{out} \) to obtain the final anomaly score map. We apply Gaussian filtering with \( \sigma = 4 \) to achieve smooth boundaries.

\begin{table*}[t!]
\centering
\small\renewcommand{\arraystretch}{1.0}

\begin{tabular*}{.99\linewidth} {@{\extracolsep{\fill}}l|l|cccccccccc}
\hline
\begin{tabular}[c]{@{}c@{}}\end{tabular} &
 \begin{tabular}[c]{@{}c@{}}\textbf{Method}\end{tabular} &
   \begin{tabular}[c]{@{}c@{}}ACC\end{tabular} &
    \begin{tabular}[c]{@{}c@{}}AI\end{tabular} &
     \begin{tabular}[c]{@{}c@{}}AMF\end{tabular} &
      \begin{tabular}[c]{@{}c@{}}ANMF\end{tabular} &
       \begin{tabular}[c]{@{}c@{}}ANI\end{tabular} &
        \begin{tabular}[c]{@{}c@{}}AP\end{tabular} &
         \begin{tabular}[c]{@{}c@{}}CS\end{tabular} &
          \begin{tabular}[c]{@{}c@{}}IS\end{tabular} &
           \begin{tabular}[c]{@{}c@{}}Mean\end{tabular}  \\ \hline
   
\multirow{5}{*}{E-b}   & \textbf{PaDiM}     & 85.5/-                      & 93.9/-             & 92.6/-             & 85.1/-             & 87.6/-             & 79.8/-                     & 87.8/-             & 76.9/-                     & 86.1/-             \\
                                    & \textbf{FastFlow}  & 77.1/-          & 82.8/-          & 68.5/-         & 81.2/-          & 78.5/-         & 52.8/-         & 71.2/- & 70.7/- & 72.8/-         \\
                                   & \textbf{RD}     & 92.2/34.3                   & 96.8/8.0           & 97.2/3.9           & 92.5/1.9           & 94.7/9.4           & 87.4/1.8                   & 93.1/5.7           & 84.8/4.7                   & 92.3/8.7           \\
                                   & \textbf{RD++}      & 91.1/32.9                   & 96.6/5.1           & 97.4/5.3           & 91.6/2.0           & 95.4/12.5          & 85.5/1.6                   & 93.1/6.8           & 84.9/4.6                   & 91.9/8.8           \\
                                   & \textbf{SimpleNet} & 75.9/13.3                   & 95.4/13.0          & 93.8/5.5           & 69.6/0.4           & 93.1/9.9           & 66.7/0.7                   & 83.4/1.9           & 81.5/5.3                   & 82.4/6.3           \\ \hline
\multirow{2}{*}{A-syn} & \textbf{DREAM}     & 63.5/20.3                   & 94.8/20.4          & 92.3/4.5           & 74.6/1.8           & 81.4/15.6          & 69.5/1.2                   & 91.0/6.2           & 76.9/4.9                   & 80.5/9.4           \\
                                   & \textbf{DeSTSeg}   & 87.8/32.5                   & 96.9/12.5          & 96.6/4.1           & 94.8/5.8           & 93.2/9.2           & 77.1/2.2                   & 90.8/3.1           & \textbf{86.9}/8.1 & 90.5/9.6           \\ \hline
R-b               & \textbf{RealNet}   & 81.0/-                      & 92.4/-             & 89.3/-             & 82.9/-             & 89.8/-             & 76.0/-                     & 81.1/-             & 78.3/-                     & 83.8/-             \\ \hline
\multirow{2}{*}{U-ni}       & \textbf{UniAD}     & 84.7/-                      & 94.6/-             & 93.5/-             & 88.0/-             & 86.0/-             & 81.4/-                     & 85.1/-             & 80.9/-                     & 86.8/-             \\
                                   & \textbf{CRAD}      & \textbf{92.9}/-    & 96.7/-             & 97.0/-             & 90.8/-             & 92.8/-             & \textbf{88.4}/-            & 91.0/-             & 86.4/-                     & 92.0/-             \\ \hline
E-b                    & \textbf{Ours}      & 91.1/\bf34.6 & \textbf{97.5/24.6} & \textbf{98.3/23.3} & \textbf{95.9/13.1} & \textbf{96.9/23.1} & 88.2/\bf2.5 & \textbf{93.5/10.2} & 85.9/\bf9.3 & \textbf{93.4/17.6} \\ \hline
\end{tabular*}%
 \caption{ 
Performance of popular IAD algorithms and our paradigm on 3CAD. We report the P-AUROC (\%) and AP (\%) metrics for each class, along with the average across all classes. Higher values indicate better performance. 
}
\label{T3}
\end{table*}

\section{Experiments}

\subsection{Experimental Settings}
\subsubsection{Datasets.}
We conduct benchmark experiments on the 3CAD and MVTec-AD datasets. The MVTec-AD dataset consists of 5,354 high-resolution images from various domains, covering 5 textures and 10 objects. The training set includes 3,629 normal images, while the test set comprises 1,725 images.

\subsubsection{Evaluation Metrics.}
Following existing methodologies, we use Area Under the Receiver Operator Curve (AUROC) and Pixel-wise Per-Region Overlap (P-PRO) for anomaly detection and localization. Specifically, I-AUROC represents image-level anomaly detection, and P-AUROC represents pixel-level anomaly localization. Additionally, we adopt Pixel-wise Average Precision (AP) to further compare the performance in detecting abnormal regions. 
\subsubsection{Implemental Details.}
All images are resized to 256 × 256. During training, we use the AdamW optimizer with a learning rate of 0.0005. The learning rate decays by a factor of 0.2 at epochs 40 and 45. We train for 50 epochs on a single NVIDIA RTX 3090 24GB with a batch size of 32.

\begin{table*}[h]
\small\renewcommand{\arraystretch}{1.0}

\begin{tabular*}{.99\linewidth} {@{\extracolsep{\fill}}l|l|cccccccccc}
\hline
\begin{tabular}[c]{@{}c@{}}\end{tabular} &
 \begin{tabular}[c]{@{}c@{}}\textbf{Method}\end{tabular} &
   \begin{tabular}[c]{@{}c@{}}ACC\end{tabular} &
    \begin{tabular}[c]{@{}c@{}}AI\end{tabular} &
     \begin{tabular}[c]{@{}c@{}}AMF\end{tabular} &
      \begin{tabular}[c]{@{}c@{}}ANMF\end{tabular} &
       \begin{tabular}[c]{@{}c@{}}ANI\end{tabular} &
        \begin{tabular}[c]{@{}c@{}}AP\end{tabular} &
         \begin{tabular}[c]{@{}c@{}}CS\end{tabular} &
          \begin{tabular}[c]{@{}c@{}}IS\end{tabular} &
           \begin{tabular}[c]{@{}c@{}}Mean\end{tabular}  \\ \hline
\multirow{5}{*}{E-b}   & \textbf{PaDiM}     & 81.6/-             & \textbf{96.1/-}    & 89.3/-             & 67.1/-             & 79.1/-             & 79.5/-             & 66.4/-    & 63.5/-    & 77.8/-             \\
                                    & \textbf{FastFlow}  & 79.1/-          & 89.5/-          & 82.3/-         & 63.0/-          & 71.9/-          & 71.3/-          & 63.4/- & 64.3/- & 73.1/-          \\
                                   & \textbf{RD}     & 90.6/82.7          & 96.0/81.5          & 89.5/82.1          & 66.8/70.0          & 81.8/81.3          & 79.0/72.1          & 74.0/77.1 & 65.9/60.8 & 80.4/75.9          \\
                                   & \textbf{RD++}      & 92.0/83.2/         & 95.5/86.5          & 89.0/84.0          & 70.0/65.7          & 81.8/85.2          & 80.5/70.8          & 76.3/\bf78.7 & 67.7/\bf63.8 & 81.6/77.2          \\
                                   & \textbf{SimpleNet} & 81.6/54.8          & 92.9/70.4          & 85.1/73.0          & 61.0/40.7          & 76.9/64.7          & 69.5/55.6          & 71.5/55.3 & 67.2/54.6 & 75.7/58.7          \\ \hline
\multirow{2}{*}{A-syn} & \textbf{DREAM}     & 80.4/-             & 89.4/-             & 73.3/-             & 61.9/-             & 78.5/-             & 71.7/-             & 68.1/-    & 70.8/-    & 74.3/-             \\
                                   & \textbf{DeSTSeg}   & 91.9/79.7          & 95.1/93.5          & 93.1/85.9          & 77.3/81.3          & 87.5/75.1             & 85.6/77.7          & 70.4/66.5 & \textbf{86.3}/60.0 & 85.9/77.4             \\ \hline
R-b               & \textbf{RealNet}   & 83.9/43.3          & 90.7/70.4          & 73.9/38.1          & 66.6/27.7          & 70.0/22.2          & 70.4/40.1          & 65.2/47.6 & 64.3/13.1 & 73.1/37.8          \\ \hline
\multirow{2}{*}{U-ni}       & \textbf{UniAD}     & 82.4               & 93.3/-             & 87.4/-             & 65.5/-             & 86.4/-             & 72.4/-             & 56.8/-    & 62.4/-    & 75.8/-             \\
                                   & \textbf{CRAD}      & 88.2/-             & 92.6/-             & 89.4/-             & 69.0/-             & 75.6/-             & 82.7/-             & 72.6/-    & 64.2/-    & 79.3/-             \\ \hline
E-b                    & \textbf{Ours}      & \textbf{93.9/84.6} & \textbf{96.1/91.8} & \textbf{94.5/90.6} & \textbf{83.8/82.4} & \textbf{90.7/88.4} & \textbf{87.2/78.5} & \textbf{77.2}/78.5 & 68.4/61.3 & \textbf{86.5/82.0} \\ \hline
\end{tabular*}%

\caption{ 
Performance of popular IAD algorithms and our paradigm on 3CAD. We report the I-AUROC (\%) and P-PRO (\%) metrics for each class, along with the average across all classes. Higher values indicate better performance. 
}
\vspace{-0.2cm}
\label{T4}
\end{table*}

\subsection{Benchmark Evaluation}
\subsubsection{Baseline Methods.}
For benchmarking and performance comparison, we select embedding-based (E-b) methods PaDiM \cite{(34)defard2021padim}, FastFlow \cite{(35)yu2021fastflow}, RD \cite{(30)deng2022anomaly}, SimpleNet \cite{(9)liu2023simplenet}, RD++ \cite{(36)tien2023revisiting}; synthesis-based (A-syn) methods DREAM \cite{(22)zavrtanik2021draem}, DeSTSeg \cite{(23)zhang2023destseg}; reconstruction-based (R-b) method RealNet \cite{(26)zhang2024realnet}; and unified (U-ni) methods UniAD \cite{(37)you2022unified}, CRAD \cite{(38)Lee2024Continuous}. We use Anomalib to reproduce PaDiM and FastFlow, while the remaining methods are implemented using their official code with original configurations.

\subsubsection{Results on 3CAD.}
The results of all methods on the 3CAD dataset are shown in Tab.~\ref{T3} and \ref{T4}. We find significant differences in performance among popular methods. Embedding-based methods, particularly teacher-student networks, perform well due to efficient feature extraction, embedding space construction, and similarity measurement, while flow models performs slightly behind. Despite the challenges posed by the deviation of synthetic anomalies from real ones, anomaly synthesis-based methods still achieve strong localization, particularly in I-AUROC and AP metrics. Combining anomaly synthesis with embedding-based approaches further improves performance. However, traditional reconstruction-based methods struggle with the challenges of the 3CAD dataset, and unified anomaly detection methods still have considerable room for improvement due to the complex distribution of each subset.

\subsubsection{CFRG on 3CAD.}
The proposed CFRG method is evaluated on the 3CAD dataset, with results shown in Tab.\ref{T3} and \ref{T4}. CFRG achieve 93.4\% AUROC, 86.5\% AUPRO, 82.0\% AP, and 17.6\%. Compared to embedding-based methods, CFRG improve P-AUROC by 1.1\% over RD, I-AUROC by 4.9\% over RD++, PRO by 4.8\%, and AP by 8.8\%. Against anomaly synthesis methods, CFRG outperform DeSTSeg in AP by 7.3\%. For reconstruction-based methods, CFRG surpass RealNet in P-AUROC and I-AUROC by 9.6\% and 13.3\%, respectively. When compared to unified anomaly detection paradigms, CFRG outperform UniAD by 6.6\% in P-AUROC and 10.7\% in I-AUROC, and CRAD by 1.4\% and 7.2\%, respectively. These results confirm the effectiveness of its coarse-to-fine localization approach over stand-alone reconstruction and distillation methods.

\begin{table}[h]
\centering
\small\renewcommand{\arraystretch}{1.0}
\resizebox{1\linewidth}{!}{
\begin{tabular}{@{}c|cccc@{}}
\hline
\multirow{2}{*}{\textbf{Methods}} & \multicolumn{4}{c}{MVTec-AD/3CAD}                  \\ \cmidrule(l){2-5} 
                                  & P-AUROC & I-AUROC & P-PRO                & AP   \\ \hline
\textbf{FastFlow}                 & 98.5/72.8    & 99.4/73.1                 & -  & -  \\

\textbf{RD++}                     & 98.2/91.9    & 99.4/81.6     & 94.977.2      & 61.5/8.8 \\
\textbf{SimpleNet}                & 98.1/82.4    & 99.6/75.7    &89.9/58.7                 & -    \\
\textbf{DREAM}                    & 97.3/80.5    & 98.0/74.3    & - & 68.4/9.4 \\
\textbf{DeSTSeg}                  & 97.9/90.5    & 98.6/85.9   & 94.7/77.4   & 75.8/9.6 \\
\textbf{RealNet}                  & 98.9/83.8    & 99.5/73.1      & 91.4/37.8           & -    \\
\textbf{UniAD}                    & 96.8/86.8    & 96.5/75.8       & -             & -    \\
\textbf{CRAD}                     & 97.8/92.0    & 99.3/79.3       & -             & -    \\
\textbf{Ours}                     & 98.4/93.4    & 98.4/86.5         & 95.6/82.0        & 73.4/17.6 \\ \hline
\end{tabular}%
}
\caption{ 
Performance comparison of popular IAD algorithms on MVTec-AD, averaged across all categories. (-) indicates unavailable metrics in the official paper.
}
\vspace{-0.2cm}
\label{T5}
\end{table}

\subsubsection{Comparison of Results.}
We compare the selected benchmarks on 3CAD and MVTec-AD, revealing several key observations, as shown in Tab.~\ref{T5}. Firstly, while CFRG performs well on 3CAD, its performance is significantly lower than on MVTec-AD. Secondly, most popular methods, which achieve around 99\% on MVTec-AD, experience a performance drop of over 10\% on our dataset. Unified-based methods also show a substantial decline due to the diversity and complexity of 3CAD, making it more challenging. Lastly, the unique complexity of 3CAD hinders existing methods, with AP metrics often below 10\% even when AUROC is high. P-PRO varies between 60\%-90\%, underscoring the increased difficulty of fine pixel-level localization in 3CAD.

\subsubsection{Qualitative Results.}
\begin{figure}
    \centering
    \includegraphics[width=.99\linewidth]{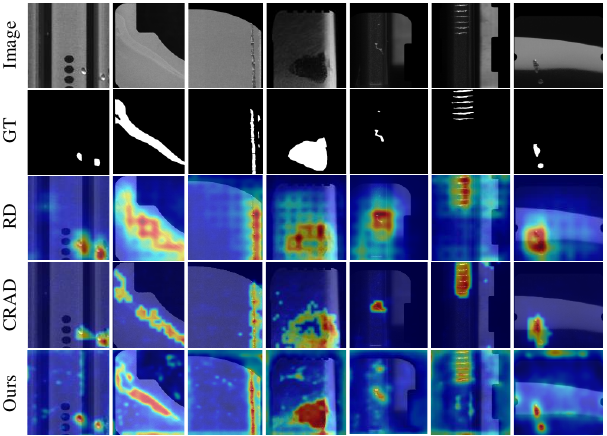}
    \caption{Qualitative illustration on 3CAD dataset.}
\vspace{-0.2cm}
\label{fig5}
\end{figure}

In Fig.~\ref{fig5}, We conduct qualitative experiments on 3CAD. Our method accurately locates defects with large morphological spans, providing finer boundaries. In densely arranged thin scratches (third and sixth columns), RD and CRAD mislocate them. In the first, fifth, and seventh columns, our method effectively pinpoints small local defects, while RD and CRAD produce wide and fuzzy localization errors. For non-structural large-area defects (second and fourth columns), our method offers more comprehensive and precise coverage compared to RD and CRAD.

\subsection{Ablation Studies}

\subsubsection{Study on Architecture Design of CFRG.}
We study the effectiveness of two key components: the recovery branch and the segmentation module. As shown in Tab.~\ref{T6}, WRC indicates the removal of the recovery branch. Retaining this branch allows the network to capture the underlying patterns of normal images when facing real anomalies, enhancing anomaly modeling. However, it may slightly interfere with the AP indicator. Overall, the combined effect is optimal, with improvements of 6.8\% on P-AUROC, 4.3\% on I-AUROC, and 22.2\% on P-PRO. WS indicates the removal of the segmentation branch. The segmentation network's ability to process local features brings significant improvements, with increases of 2.8\% on P-AUROC, 3.1\% on I-AUROC, 7.2\% on P-PRO, and 9.1\% on AP.

\subsubsection{Study on Distillation Paradigms.}
The feature distribution of the distillation process significantly impact localization performance, as shown in Tab.~\ref{T6}.WP denotes that we removed the practice of pushing the abnormal feature distance between the teacher and student networks, focusing solely on closing the normal feature distance. This led to decreases of 0.1\%, 0.6\%, 0.5\%, and 1.1\% across the four metrics. When dealing with weak defects, the introduction of this push-pull paradigm aids in separating abnormal areas. This approach aligns with our goal of using heterogeneous teacher and student networks.
\subsubsection{Study on Recovery guidence.}
We further examined the impact of the hint weight from the recovery branch. In the WC scenario, we used only \(in_t\) as the input to the segmentation network, removing the auxiliary weight from the recovery branch. This led to performance drops of 0.3\%, 0.6\%, 0.7\%, and 1.5\% across the four metrics. This demonstrates that the recovery branch not only enhances the final prediction but also helps reduce distribution deviation introduced during the distillation stage.
\subsubsection{Study on Heterogeneous.}
We use heterogeneous teacher and student networks to ensure that weak defect regions have distinct feature representations, promoting feature separation and accelerating network optimization during distillation. In the WH scenario, we replaced the student network with WideResNet50, the same architecture as the teacher. This change resulted in performance drops of 0.5\%, 0.8\%, 1.1\%, and 0.5\% across the four metrics. The improvement achieved by using a student network with a different architecture highlights the potential for further research within the heterogeneous paradigm.

\begin{table}[t]
\centering
\small\renewcommand{\arraystretch}{1.0}
\begin{tabular}{c|cccccc}
\toprule
\textbf{WRC}   &   & \checkmark  & \checkmark  & \checkmark & \checkmark    & \checkmark \\
\textbf{WS}     & \checkmark    &  & \checkmark & \checkmark & \checkmark   & \checkmark \\
\textbf{WP}     & \checkmark      & \checkmark      &  & \checkmark & \checkmark    & \checkmark\\
\textbf{WC}     & \checkmark        & \checkmark         & \checkmark          &  & \checkmark  & \checkmark\\
\textbf{WH}   & \checkmark         & \checkmark          & \checkmark          & \checkmark  &     & \checkmark      \\
\midrule
\textbf{P-AUROC$\uparrow$} & 86.6      & 90.6      & 93.3      & 93.0      & 92.9      & \bf93.4      \\
\textbf{I-AUROC$\uparrow$} & 82.2      & 79.1      & 85.9      & 85.9      & 85.7      & \bf86.5      \\ 
\textbf{P-PRO$\uparrow$} & 59.8      & 74.8      & 81.5      & 81.3      & 80.9      & \bf82.0      \\ 
\textbf{AP$\uparrow$} & \bf19.7      & 8.5      & 16.5      & 16.1      & 17.1      & 17.6      \\ 

\bottomrule
\end{tabular}
\caption{ 
Ablation studies of CFRG with AUROC, P-PRO, and AP metrics.
}
\vspace{-0.2cm}
\label{T6}
\end{table}

\section{Conclusion}
Given the current state of industrial anomaly detection, this paper introduces 3CAD, the first large-scale dataset focused on anomaly detection in the manufacturing process of 3C product parts. Derived from real production lines, 3CAD provides pixel-level annotations for eight different types of industrial product parts, comprising 27,039 high-resolution images. The samples and defects offer a realistic representation of manufacturing scenarios. 
To tackle the challenges of 3CAD, we propose CFRG, a method that uses heterogeneous teacher-student networks to generate coarse localization, which is then refined through a segmentation network with recovery guidence. Our approach demonstrates excellent performance. Future work will focus on developing more advanced localization paradigms to further enhance accuracy.

\section{Acknowledgments}
This work was supported in part by the National Natural Science Foundation of China (Grant No. 62372284).

\bibliography{ref}
\newpage
\appendix

\section*{Supplementary Material}

In this supplementary material, we first provide details on the dataset construction, including the setup of data acquisition equipment and preprocessing steps applied to the labeled data, such as label format verification and data trimming. Next, we present visualizations of the 3CAD dataset, showcasing normal images in various forms for each product, as well as visualizations of different products or instance locations for each defect. Finally, we offer detailed experimental metrics for the CFRG paradigm on the MVTec AD dataset, along with a study on the impact of loss weights.
\begin{figure*}[h!]
    \centering
    \includegraphics[width=.99\linewidth]{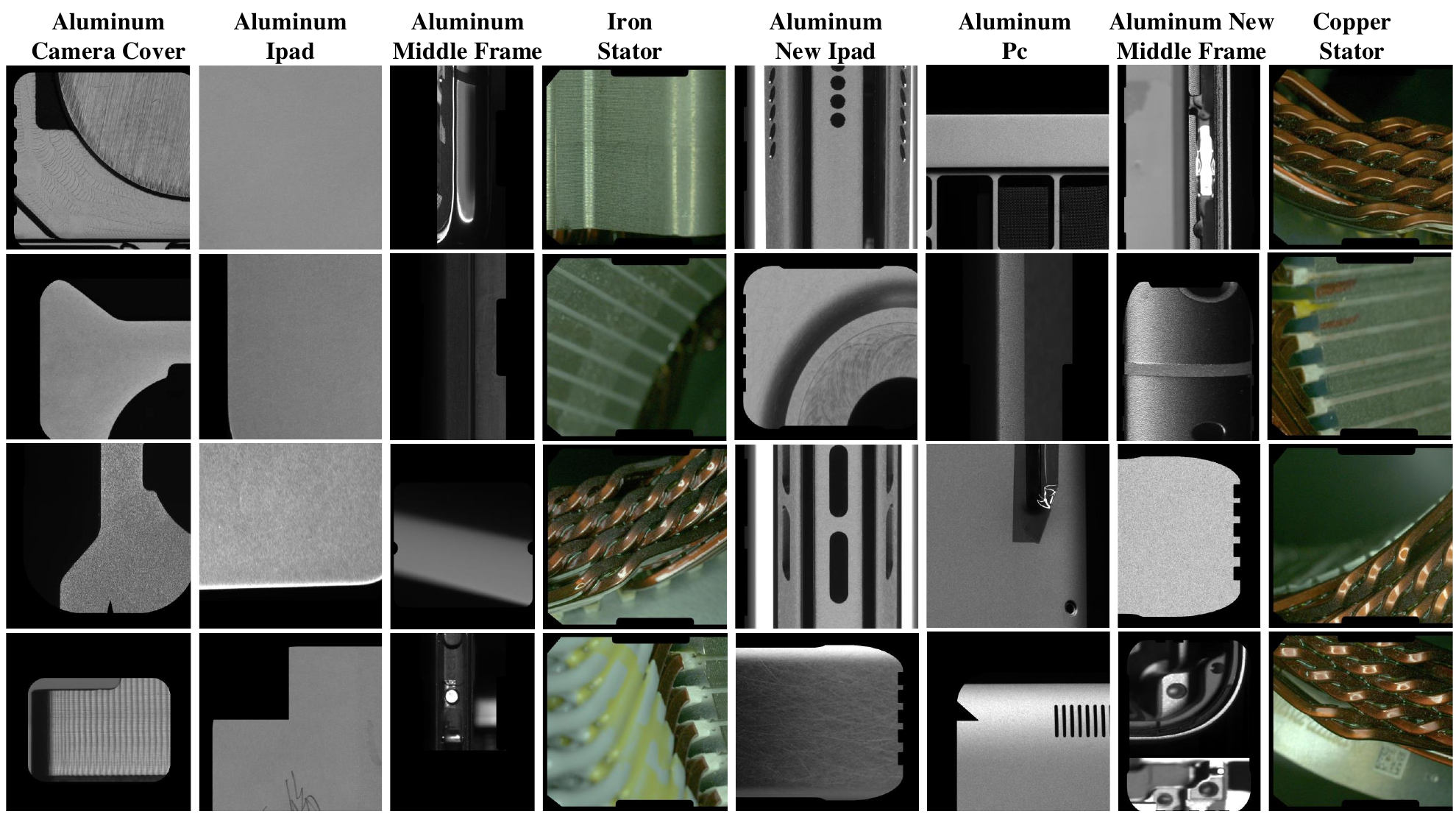}
    \caption{Normal visualization in 3CAD: Each column represents a product, and each row displays a different industrial image of that same product.}
\label{fig6}
\end{figure*}
\section{Dataset Construction Detail}

\myparagraph{Prototype Construction}
To ensure optimal defect detection, the assembly device structure and light source are customized according to the characteristics and the frequency of specific defect types of product. For instance, when collecting mobile phone camera cover, a line scan camera is employed. The distance and angle of camera are precisely adjusted, and the device structure is assembled to capture images via top-line scanning. Each product is photographed twice, with a 90-degree rotation between shots. During this process, the product must move at a constant speed to prevent any motion blur or tilting, and the height of the two products on the same robot must remain consistent to ensure identical positioning for each scan. To enhance defect detection accuracy, a blue strip light is used to illuminate the product at a fixed angle and distance. The camera model used is the MV-CL041-70GM, with a resolution of 4096.

\myparagraph{Data preprocessing}
After completing the data annotation, we conducted an initial review to identify and correct any labeling errors. Next, we examined the annotations to rectify invalid polygons, such as those with only two points or intersecting vertices. Given the large and inconsistent resolutions of the original images, we used the SAHI library to crop the images, ensuring that the maximum resolution is maintained at 1024 pixels, while images with resolutions below 1024 pixels are kept at their original size. The cropping process is performed by sliding from left to right and top to bottom. To better preserve the original defect areas, especially since small defect samples constitute a significant portion of the data, we applied a 20\% overlap between the sliding windows. Images with dimensions less than 20\% of the window size were discarded. Most existing anomaly detection methods rescale images to 256 or 224 pixels, and since small defects are prevalent in our dataset, this approach helps mitigate information loss during rescaling. Additionally, this makes the data more accessible for researchers.

\myparagraph{Benchmark Evaluation Implemental Details}
We selected ten popular anomaly detection algorithms for benchmarking. For Padim and Fastflow, we used the default settings in Anomalib, scaling all images to 256 × 256 pixels. For the other methods, we followed the official configurations. Images were scaled to 256 × 256 pixels for DREAM, DeSTSeg, RD++, RD, and RealNet, while UniAD, SimpleNet, and CRAD images were scaled to 224 × 224 pixels. All other parameters, such as learning rate and batch size, were kept as per the official implementations.

\section{3CAD Visualisation}

\myparagraph{Visualization of each product}
In Fig.~\ref{fig6}, we present the visualization for each product, highlighting their distinct morphological features, which arise from differences in part structure and the optical surfaces used during imaging. The figure also reveals the complex geometric structures of the products, with even the same product exhibiting varying feature distributions.

\begin{figure*}[h]
    \centering
    \includegraphics[width=.99\linewidth]{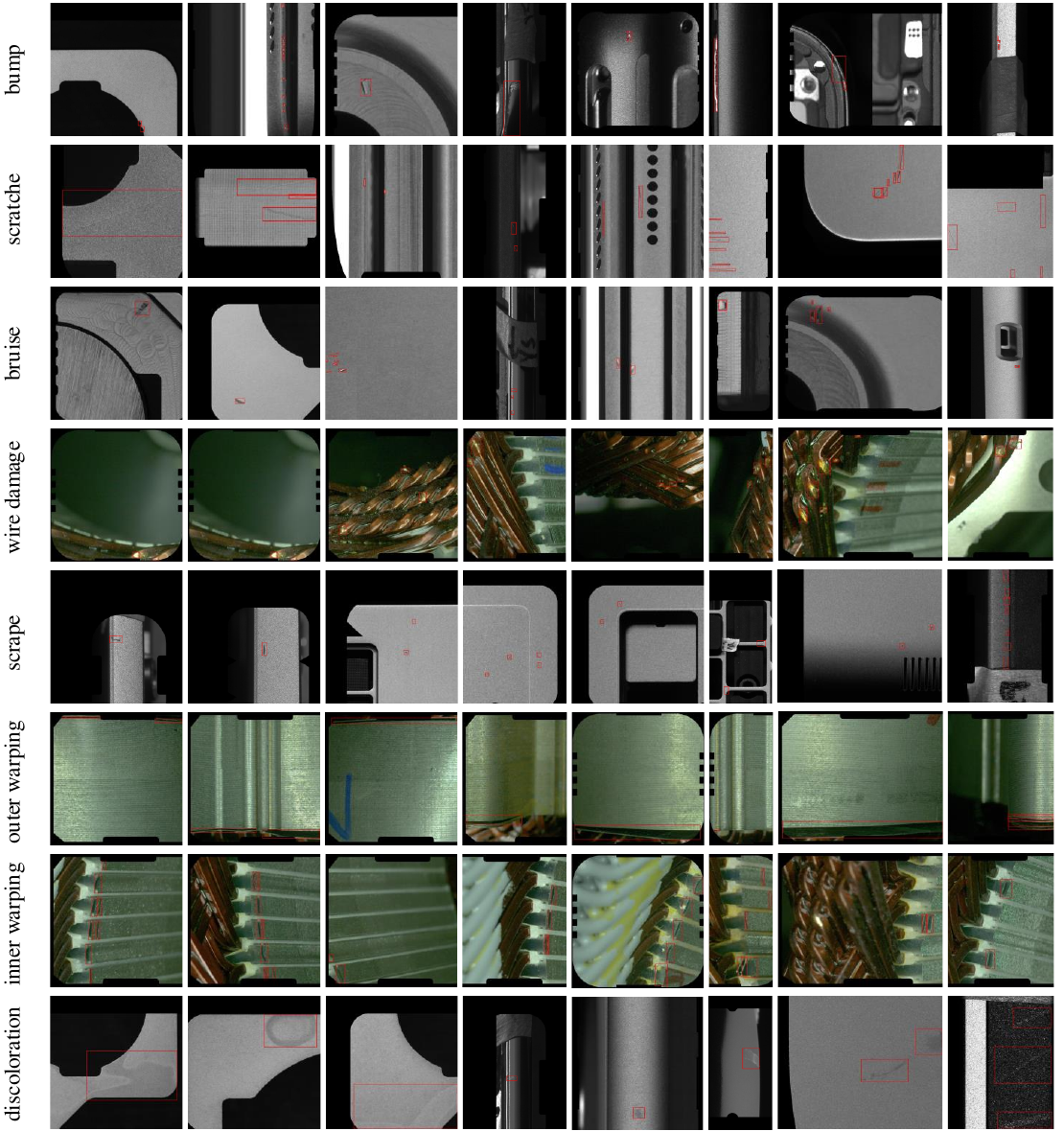}
    \caption{Anomaly visualization in 3CAD: Each row represents a different anomaly category across various products and locations. From top to bottom, the categories are bump, scratch, bruise, upper line, scrape, outer warping, inner warping, and discoloration, respectively.}
\label{fig7}
\end{figure*}

\begin{figure*}[t]
    \centering
    \includegraphics[width=.99\linewidth]{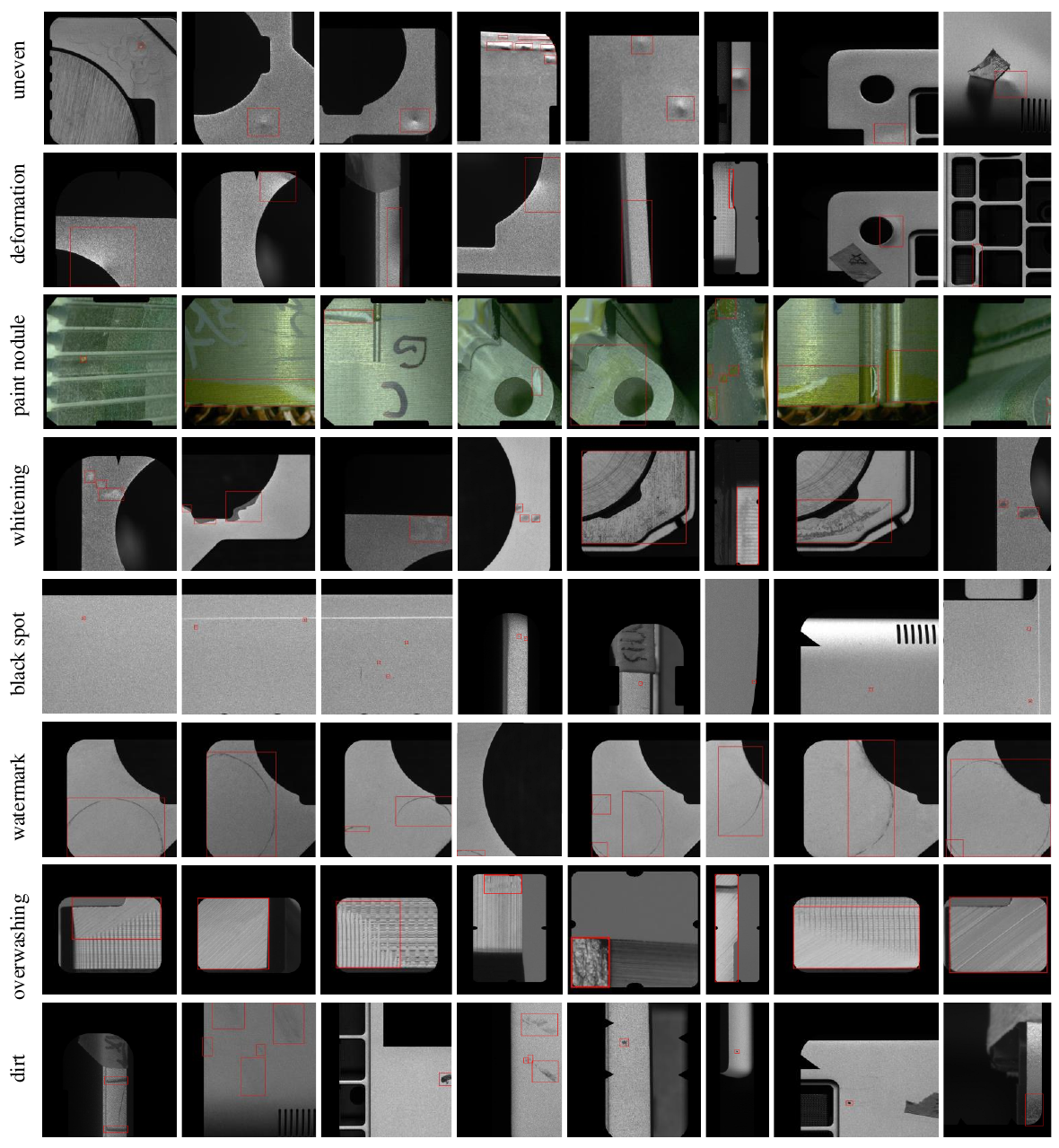}
    \caption{Anomaly visualization in 3CAD: Each row represents a different anomaly category across various products and locations. From top to bottom, the categories are uneven, deformation, iron core airflow, whitening, black spots, watermarks, overwashing, dirt, respectively.}
\label{fig8}
\end{figure*}

\begin{figure*}[t]
    \centering
    \includegraphics[width=.99\linewidth]{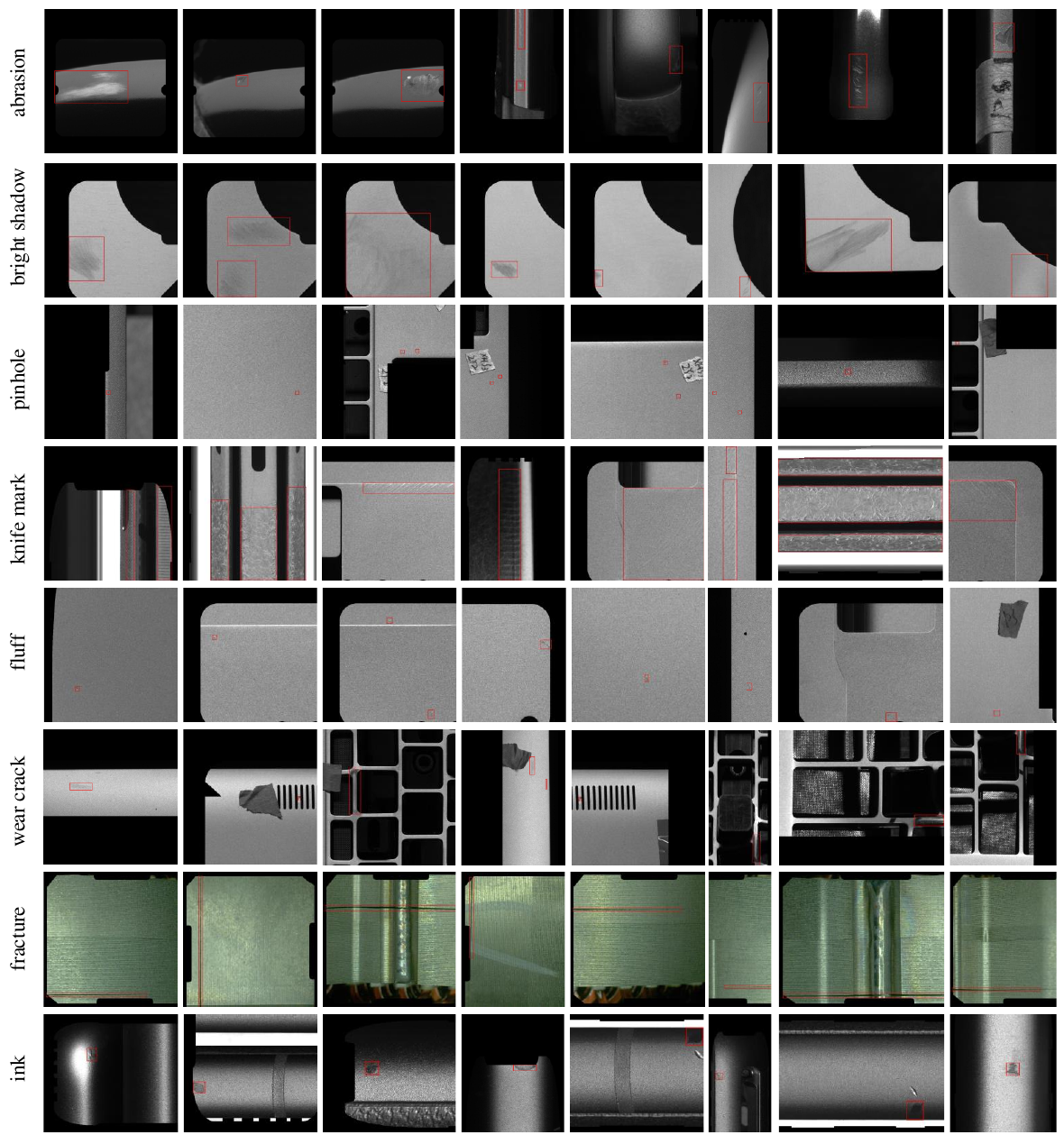}
    \caption{Anomaly visualization in 3CAD: Each row represents a different anomaly category across various products and locations. From top to bottom, the categories are abrasion, bright shadow, pinhole, knife mark, fluff, wear crack, fracture, and ink, respectively.}
\label{fig9}
\end{figure*}

\myparagraph{Visualization of each defect type}
Here, we present a more detailed data visualization, showcasing abnormal images representing 24 defect types, including bump, scratche, bruise, wire damage, scrape, outer warping, inner warping, discoloration, uneven, deformation, paint nodule, whitening, black spot, watermark, overwashing, dirt, abrasion, bright shadow, pinhole, knife mark, fluff, wear crack, fracture, and ink, as shown in Fig.~\ref{fig7}, ~\ref{fig8}, and ~\ref{fig9}. The 3CAD dataset encompasses a wide range of both common and uncommon defect types, totaling 24 unique defects, which present real-world challenges. First, even within the same defect category, there can be considerable variation in form, whether in size, shape, or visual characteristics. Second, defects are often difficult to detect due to their similarity to the background, their small size, and the presence of noise, all of which complicate defect localization. Third, because defect formation is influenced by environmental factors and production techniques, defects can appear in any location and in various forms, leading to issues with similarity interference between different defect types. For instance, when dirt and discoloration in an image are both black or white and have similar shapes, distinguishing between them becomes challenging. In practice, dirt can be removed with wiping, whereas discoloration cannot.

\begin{table*}[h!]
\centering
\scriptsize\renewcommand{\arraystretch}{1.0}
\resizebox{1\linewidth}{!}{%
\begin{tabular}{c|cccccccccc|c}
\hline
\multirow{2}{*}{Metrics} & \multicolumn{10}{c|}{Objects}                                                                  & \multirow{2}{*}{Mean} \\ \cmidrule(l){2-11}
                         & Bottle & Cable & Capsule & Hazelnut & Metal nut & Pill & Screw & Toothbrush & Transistor & Zipper &                       \\ \cmidrule(l){1-12}
P-AUROC                  & 99.3   & 98.1  & 98.7    & 99.0     & 98.0      & 97.4 & 99.1  & 99.2       & 92.5       & 99.3   & 98.0                 \\
I-AUROC                  & 100.0  & 95.2  & 96.7    & 99.7     & 100.0     & 97.5 & 90.9  & 98.3       & 98.2       & 99.6   & 97.6                 \\
P-PRO                    & 98.2   & 93.3  & 94.4    & 95.0     & 94.7      & 96.7 & 96.0  & 94.8       & 83.0       & 98.0   & 94.4                 \\
AP                       & 91.0   & 71.7  & 61.2    & 74.0     & 85.0      & 79.4 & 48.3  & 63.6       & 57.2       & 79.2   & 71.0                 \\ \hline            
\end{tabular}%
}
\caption{ 
The detailed metrics of our method on the MVTec AD dataset demonstrate that CFRG also delivers strong performance on widely used datasets. 
}
\label{T7}
\end{table*}

\begin{table}[h!]
\centering
\scriptsize\renewcommand{\arraystretch}{1.0}
\resizebox{1\linewidth}{!}{%
\begin{tabular}{c|ccccc|c}
\hline
\multirow{2}{*}{Metrics} & \multicolumn{5}{c|}{Textures} & \multirow{2}{*}{Mean} \\ \cmidrule(l){2-6}
                         & Carpet & Grid  & Leather & Tile  & Wood &       \\ \cmidrule(l){1-7}
P-AUROC                  & 99.5   & 99.5  & 99.8    & 99.1  & 97.4 & 99.0 \\
I-AUROC                  & 99.9   & 100.0 & 100.0   & 100.0 & 99.4 & 99.8 \\
P-PRO                    & 98.5   & 98.0  & 99.5    & 97.2  & 96.5 & 97.9 \\
AP                       & 83.1   & 57.4  & 78.1    & 93.1  & 78.6 & 78.0 \\ \hline            
\end{tabular}%
}
\caption{ 
The detailed metrics of our method on the MVTec AD dataset demonstrate that CFRG also delivers strong performance on widely used datasets. 
}
\label{T8}
\end{table}


\begin{table}
    
\centering
\small\renewcommand{\arraystretch}{1.0}

\resizebox{1\linewidth}{!}{
\begin{tabular}{@{}ccccccc@{}}
\toprule
$L_{dis}$                  & $L_{rec}$                  & \multicolumn{1}{c|}{$L_{bce}$} & P-AUROC$\uparrow$ & I-AUROC$\uparrow$ & P-PRO$\uparrow$ & AP$\uparrow$ \\ \midrule
1/3                   & 1/3                   & \multicolumn{1}{c|}{1/3}  & 93.3              & 85.2              & 81.4            & 17.0         \\
1                     & 1/10                  & \multicolumn{1}{c|}{1}    & 92.7              & 85.7              & 80.4            & 17.6         \\
1/10                  & 1                     & \multicolumn{1}{c|}{1}    & 93.3              & 85.8              & 81.6            & 17.3         \\
\multicolumn{1}{l}{1} & \multicolumn{1}{c}{1} & \multicolumn{1}{c|}{1/10}  & 93.1              & 85.9              & 81.7            & 14.2         \\
1                     & 1                     & \multicolumn{1}{c|}{1}    & 93.4              & 86.5              & 82.0            & 17.6         \\ \bottomrule
\end{tabular}%
}
\caption{ 
Studies of Loss weight with AUROC, P-PRO, and AP metrics.
}
\label{T9}
\end{table}

\section{Supplementary Experimental Results}
\myparagraph{Result on MVTec AD}
As shown in Tab.~\ref{T7} and Tab.~\ref{T8}, we present the detailed metrics for CFRG across each category on the MVTec AD dataset, including P-AUROC, I-AUROC, P-PRO, and AP. Our method achieves results close to the state-of-the-art on mainstream datasets, demonstrating strong performance. Notably, while the metrics show minimal variation on MVTec AD, there is a significant performance improvement on 3CAD. This indicates that our method is particularly effective in addressing the challenges posed by small and subtle defects. 

\myparagraph{Study on Loss Weight}
As shown in Tab.~\ref{T9}, varying the weights of the distillation loss (\(L_{\text{dis}}\)), recovery loss (\(L_{\text{rec}}\)), and segmentation loss (\(L_{\text{bce}}\)) impacts performance. Reducing any loss weight leads to decreased performance metrics. For instance, lowering \(L_{\text{rec}}\) to 1/10 decreases P-AUROC to 92.7\% and P-PRO to 80.4\%, though I-AUROC and AP slightly improve. Additionally, the first row demonstrates that the magnitude of each loss component is also critical, even when the weights are equal. Thus, both balancing the weights of all losses and maintaining their magnitude are essential for achieving optimal anomaly detection performance.

\end{document}